\pdfoutput=1
\documentclass[11pt, table]{article}

\usepackage[final]{acl}

\usepackage{times}
\usepackage{latexsym}

\usepackage[T1]{fontenc}

\usepackage[utf8]{inputenc}

\usepackage{microtype}

\usepackage{inconsolata}

\usepackage{graphicx}

\usepackage{makecell}

\usepackage{multirow}

\usepackage{subcaption}

\title{From Isolates to Families: Using Neural Networks for Automated Language Affiliation}

\author{
 \textbf{Frederic Blum\textsuperscript{1,2}},
 \textbf{Steffen Herbold\textsuperscript{3}},
 \textbf{Johann-Mattis List\textsuperscript{1,2}} \\
 \textsuperscript{1}Department of Linguistic and Cultural Evolution, Max-Planck Institute for Evolutionary Anthropology, 04103 Leipzig, \\
 \textsuperscript{2}University of Passau, Chair of Multilingual Computational Linguistics, Passau, 94032, Germany, \\
 \textsuperscript{3}University of Passau, Chair of AI Engineering, Passau, 94032, Germany \\
 \small{\textbf{Correspondence:} \href{mailto:frederic_blum@eva.mpg.de}{frederic\_blum@eva.mpg.de}}
}
\begin{document}
\maketitle
\begin{abstract}
In historical linguistics, the affiliation of languages to a common language family is traditionally carried out using a complex workflow that relies on manually comparing individual languages. Large-scale standardized collections of multilingual wordlists and grammatical language structures might help to improve this and open new avenues for developing automated language affiliation workflows.
Here, we present neural network models that use lexical and grammatical data from a worldwide sample of more than 1,000 languages with known affiliations to classify individual languages into families.
In line with the traditional assumption of most linguists, our results show that models trained on lexical data alone outperform models solely based on grammatical data, whereas combining both types of data yields even better performance.
In additional experiments, we show how our models can identify long-ranging relations between entire subgroups, how they can be employed to investigate potential relatives of linguistic isolates, and how they can help us to obtain first hints on the affiliation of so far unaffiliated languages. 
We conclude that models for automated language affiliation trained on lexical and grammatical data provide comparative linguists with a valuable tool for evaluating hypotheses about deep and unknown language relations.
\end{abstract}

\section{Introduction}
One of the central tasks in historical linguistics is the grouping of languages into families. While this is often done to propose new language families, this task also includes the affiliation of individual languages into existing families. The affiliation of languages to a common language family or a subgroup within that family is usually carried out manually and relies on comparing individual languages in depth.

The traditional `comparative method' in historical linguistics uses lexical and grammatical features to assign individual languages to one of the more than 200 language families proposed so far \cite{Osthoff1878, Anttila1972, Durie1996}. The main part of this comparison starts with the initial assumption that two languages are related. However, such family relationships do not come as given and must first be established \cite{Hoenigswald1978, Nichols1996, Donohue2012, Campbell2017}. While the rise of computational methods in historical linguistics has largely replaced traditional techniques for subgrouping with computational approaches in phylogenetic reconstruction \cite{Greenhill2020, Wu2020, Blum2023}, the classical workflow for language \emph{affiliation} of the comparative method is still considered the state-of-the-art in the field.
But given that the affiliation of languages to families can be viewed as a computational \emph{classification task}, it is possible to model the task in a setting that benefits from a large number of digital cross-linguistic datasets that have recently been published \citep{List2022e, DoReCo-1.2, ASJPv20, Skirgaard2023, Blum2024}.

We show how \emph{automated language affiliation} can be implemented computationally, how this approach can recover known long-distance genealogical relationships, and how it can shed new light on the affiliation of linguistic isolates and small language families. We train models on known language families and affiliate languages of unknown classification with the existing ones while aiming for worldwide coverage. When reporting the results, we pay special attention to the success of classifying small language families. The supervised training approach enables us to build upon the vast existing knowledge of historical linguistics. Having a large baseline of classes corresponding to true language families makes it possible to go beyond individual comparisons and to compare against all families simultaneously.

\section{Background}
Starting with the detection of the Indo-European \citep{Bopp1816} and Uralic \citep{Gyarmathi1799} families towards the beginning of the 19th century, linguists have been able to group the more than 7000 languages still spoken today into several hundred language families. However, while major parts of the traditional workflow of the comparative method have been intensively discussed and partially formalized, the first step of this workflow, the \emph{affiliation} of languages to a family by \emph{proving} their genetic relationship \citep[``proof of relationship'', see][]{Durie1996} still lacks formalization. 
Although many quantitative and qualitative approaches have been proposed throughout the 20th and 21st centuries, none except the comparative method has gained general acceptance.
Some methods, such as the application of superficial ``mass comparison'' techniques to large wordlists \cite{Greenberg1957}, have been heavily criticized because of their lack of standardized data and clear criteria, and ultimately fell out of vogue \citep{Campbell1988}. Other methods, such as the proposal by \citet{Dolgopolsky1964}, who suggested looking for matching consonant classes to identify potential cognates, were ignored by most scholars for a long time before they were re-adopted in different contexts.
 
While scholars working in the traditional paradigm of the comparative method usually agree that lexical and grammatical evidence combined is best to prove language relationship \citep{Campbell2008}, the former is given preference in those cases where grammatical evidence is hard to obtain \citep{Dybo2008}. Quantitative and statistical methods typically restrict themselves to either lexical \emph{or} grammatical evidence.

Quantitative methods that take lexical data as their primary source can be divided into two basic types, depending on the evidence they try to obtain. Some approaches concentrate on the regularity of sound correspondences, trying to show that pairs of related languages exhibit significantly more matches in sound correspondences than unrelated ones \citep{Ringe1992, Kessler2001, Blevins2021}. Other methods do not use specific sounds as observed in the languages in question and convert them to broader classes (\emph{sound classes} or \emph{consonant classes}, as originally proposed by \citealt{Dolgopolsky1964}, see \citealt{List2014d}) to identify cognate words. These methods argue that words that share direct matches in a certain number of sound classes are likely to be etymologically related and that languages for which a certain number of matches can be observed are likely to be genetically related \citep{Baxter2000, Turchin2010, Kassian2023}. Among the latter approaches, the \emph{Automated Similarity Judgment Program} (ASJP, \href{https://asjp.clld.org}{https://asjp.clld.org}, \citealt{ASJPv20}) deserves special mention, given that it can be seen as a first attempt to automatically classify as many of the languages of the world as possible with the help of phylogenetic methods. Using a specific sound class alphabet by which speech sounds are reduced to 40 classes, ASJP computationally compares word forms in language pairs, using traditional methods for sequence alignment \citep{Wagner1974}, to infer distances between language pairs. These are later used to reconstruct a phylogenetic tree with the help of the \emph{Neighbor-joining} algorithm \citep{Saitou1987}. 

Quantitative methods that exclusively use grammatical data as their primary evidence have less frequently been proposed than their lexical counterparts. Despite this, \citet{Dunn2005} suggest that analyzing grammatical data could lead further back in time than the traditional comparative method. Today, these claims have lost supporters due to other studies indicating that grammatical features alone are less well suited for language classification because they diffuse easily in cases of language contact \cite{Gray2010, Greenhill2010}. The high potential for such diffusion is due to the limited amount of variation that grammatical features exhibit \cite{Wichmann2017}. However, these dynamics remain understudied, and we lack further case studies to analyze the behavior of grammatical data in large-scale classification settings.

We can now refine those early automated classification methods thanks to the release of new databases \cite{List2022e, Skirgaard2023}. 
The key difference between automated language affiliation and previous computational approaches to language classification is adopting a supervised learning approach. Our method directly benefits from previously established classifications based on the comparative method. It allows us to affiliate previously unclassified languages to existing language families, therefore strictly following an incremental approach to language classification. This approach also allows us to test hypotheses of deep language families, as we will show in our case studies, or to re-consider the affiliation of language isolates to other language families.

We test the model predictions in three case studies (Indo-European, Sino-Tibetan, and Uto-Aztecan) to evaluate the model classification on established language families sharing a long common history. Further, we test the affiliation of four language isolates: Basque, Bangime, Kusunda, and Mapudungun. We also show how this method can contribute to affiliating historical data of unknown classification to existing language families. At the same time, we preserve a conservative approach by including linguistic isolates in the training to restrain the model from unsubstantiated speculation in the form of false positives. 

\section{Materials and Methods}
\subsection{Cross-Linguistic Data}
We use Lexibank \citep[v2.0,][]{Blum2025} and Grambank \citep[v1.0.3,][]{Skirgaard2023}, the two currently largest collections of standardized lexical and grammatical data, to train our model. Both databases are created and published using the \emph{Cross-Linguistic Data Formats} (CLDF, \citealt{Forkel2018}, \href{https://cldf.clld.org}{https://cldf.clld.org}), in which common linguistic constructs, such as \emph{language}, \emph{concept}, and \emph{sound}, are linked to reference catalogs, such as Glottolog \citep[\href{https://glottolog.org}{https://glottolog.org},][]{Glottolog} for languages, Concepticon \citep[\href{https://concepticon.clld.org}{https://concepticon.clld.org},][]{Concepticon} for concepts, and CLTS \citep[\href{https://clts.clld.org}{https://clts.clld.org},][]{CLTS} for speech sounds. This ensures the standardization and comparability of data both within and across datasets. 
To test the quality of the data provided by Lexibank, we train an additional lexical model using the ASJP data \citep[v20,][]{ASJPv20}, which is also available in CLDF.


\subsection{Data Vectorization}
\begin{table*}[th]
    \resizebox{\textwidth}{!}{%
    \tabular{cc}
        \begin{tabular}[t]{|l|l|l|l|}
        \hline
            \bfseries Segments &\bfseries  Sound Cl. &\bfseries  Cons. Cl. &\bfseries  Vector  \\ \hline\hline
            d e & \ttfamily TV & \ttfamily T~- & \makecell[l]{\ttfamily {[}1000000000{]} \\ \ttfamily {[}0000000000{]}} \\ \hline
            n a l a &\ttfamily  NVRV &\ttfamily  NR & \makecell[l]{\ttfamily {[}0100000000{]} \\ \ttfamily {[}0010000000{]}} \\ \hline
            n e r a &\ttfamily  NVRV &\ttfamily  NR & \makecell[l]{\ttfamily {[}0100000000{]} \\ \ttfamily {[}0010000000{]}} \\ \hline
            k o k o n &\ttfamily  KVKVN &\ttfamily KK & \makecell[l]{\ttfamily {[}0001000000{]} \\ \ttfamily {[}0000100000{]}} \\ \hline
        \end{tabular} &
        \begin{tabular}[t]{|l|l|c|c|}
            \hline
                \bfseries  Parameter &\bfseries  Type &\bfseries  Value &\bfseries  Vector  \\ \hline\hline
                Fixed order S/A/P & Bin & 0 & \ttfamily [10] \\ \hline
                Fixed order S/A/P & Bin & 1 & \ttfamily [01] \\ \hline
                Fixed order S/A/P & Bin & - &\ttfamily  [00] \\ \hline 
                \multirow{4}*{Order of NUM and N} & 1-3 & 1 & \ttfamily [10] \\
                 & 1-3 & 2 & \ttfamily [01] \\
                & 1-3 & 3 & \ttfamily [11] \\
                & 1-3 & - & \ttfamily [00] \\ \hline
            \end{tabular}
\endtabular}
    \caption{Vectorization of words and grammatical features into vectors. The left table presents the conversion of segments in lexical forms into Dolgopolsky classes. Each class is assigned an index value in the vector, and the corresponding index set to 1. The same procedure applies to the binary Grambank features in the right table. In some cases, the value `3' represents the meaning `both orders are attested', hence both vector indices are set to `1'.}
    \label{tab:lb_vector}
\end{table*}

\subsubsection{Lexibank and ASJP}
For Lexibank, we use the 100 concepts that are part of the Swadesh-100 list \citep{Swadesh1955}, given that this list has sufficient coverage across the dataset. However, the method can be used with any concept list that has been standardized in Concepticon \citep{Concepticon}. For ASJP, we use their original 40-item wordlist \cite{Holman2008}.

We convert the lexical forms into vectors that can be used as input for the neural network. This processing step is illustrated in Table~\ref{tab:lb_vector}. Each segment in Lexibank and ASJP is standardized phonetically. To improve the comparability, we convert the sounds to their corresponding \emph{Dolgopolsky class}, based on the sound classes proposed by \citet{Dolgopolsky1964}. Those ten classes are based on the likelihood of sound change between individual sounds. Such sound change happens frequently between sounds within a class of similar articulation, but not between sounds of different classes \citep[for details, see][]{List2014d}. Following the original proposal, only the first two consonant classes are considered. All additional consonants as well as vowels -- except for word-initial vowels, which are considered a special consonant class \textsc{h} -- are removed from the representation. We then convert the classes to indices of a one-hot vector of length 10, the number of Dolgopolsky sound classes, with the index of the attested class set to 1. All vector indices remain zero if a word is not attested in the language. All individual word vectors are concatenated for the language, and given as input to the model.

\subsubsection{Grambank}
Grambank is a database of 195 typological features from more than 2,000 languages \cite{Skirgaard2023}. Since most features are coded as binary, converting them into vectors is more straightforward than with Lexibank. Each value is mapped to an index in a one-hot vector of length two, and the index related to the attested value in a language is set to 1. Some word order features have three possible values, where `3' represents the meaning `both 1 and 2 are attested'. Therefore, both indices in the vector are set to 1.

\subsection{Baseline}
We designed a fast and simple test to have a baseline comparison for our neural network models, based on the idea that the number of matching consonant classes can give hints on cognate words \cite{Dolgopolsky1964, Turchin2010}. In this baseline, we compare one language with unknown affiliation $L_1$ against all the other language varieties in the training data and assign it to the language family of the language $L_{max}$ that shares the largest number of words matching in the two first consonant classes with $L_1$.

\subsection{Neural Network Model Training}
The neural network models are trained on the input vector of lexical, grammatical, or combined data as well as the labels of the language families. We train the models on an 80\% sample of the data stratified by family labels and evaluate the performance of each model on the remaining 20\% to find the best-performing one based on the balanced accuracy across language families \cite{Brodersen2010}. This split is necessary to avoid over-fitting the model to the training data \cite{dietterich1995overfitting}. Even though a split into an additional development- or tuning-set would improve the model training even further \cite{van-der-goot-2021}, the data scarcity makes this a difficult enterprise. To test the robustness of our results despite the small sample size, we run each model 100 times with random seeds, so that different train/test sets are used \cite{Gorman2019, Vabalas2019, Coltekin2020}. As an additional measure against over-fitting, we implement an early-stopping strategy during training \cite{Ying2019}. 

\subsection{Evaluation}
We conduct four different experiments. The first experiment (§~\ref{test1}) consists of a model comparison using a common subset of languages attested in ASJP, Lexibank, and Grambank. This comparison tests our baseline and the neural network models using an identical selection of languages.

In the second experiment (§~\ref{test2}) we evaluate how well our models can affiliate entire subgroups with the correct language family. We select three language families -- Indo-European, Sino-Tibetan, and Uto-Aztecan -- in which larger branches have split off considerably early during their evolution. We then train our models without the languages corresponding to these branches and test how automated language affiliation succeeds in assigning these languages to the correct family.

In the third experiment (§~\ref{test3}), we investigate how our models affiliate language isolates, that is, languages which could so far not been convincingly assigned to \emph{any} established language family. We use \href{https://glottolog.org/resource/languoid/id/bang1363}{Bangime}, \href{https://glottolog.org/resource/languoid/id/basq1248}{Basque}, \href{https://glottolog.org/resource/languoid/id/kusu1250}{Kusunda}, and \href{https://glottolog.org/resource/languoid/id/mapu1245}{Mapudungun} as exemplary case studies.

In the fourth and final experiment (§~\ref{test4}), we demonstrate with the example of Cararí \cite{Natterer} how historical languages that have not been affiliated with any language family so far can be investigated with our automated language affiliation models.

\subsection{Implementation}
We implemented our models with PyTorch (v2.5.1, \citealt{Ansel2024}) in a feed-forward neural network with two hidden layers with a ReLU activation function. The hidden layers have a size of four times the number of language families. We process the datasets with SQLite (\href{https://www.sqlite.org/}{https://www.sqlite.org/}) after conversion from CLDF via PyCLDF (v1.40.4, \citealt{PyCLDF}). We converted the segments to sound classes with LingPy (v2.6.13, \citealt{LingPy}).

We use a weighted CrossEntropy loss function to better account for the many small language families present in our data \cite{Zhang2018}. We used an Adam optimizer with a learning rate of 1e-3. The batch size we used is 2048. The hyperparameters were chosen on individual model comparisons between common values. Each training run consists of 5,000 epochs and is canceled if no improvement is made for 500 epochs. The models were trained on a V100 GPU node on a high-performance cluster, taking approximately 90 minutes. They can also be trained on ordinary computers due to the small size of the underlying data.

\section{Results}
\subsection{Initial Model Comparison}\label{test1}
\begin{figure*}[t]
\centering
        \includegraphics[width=\textwidth]{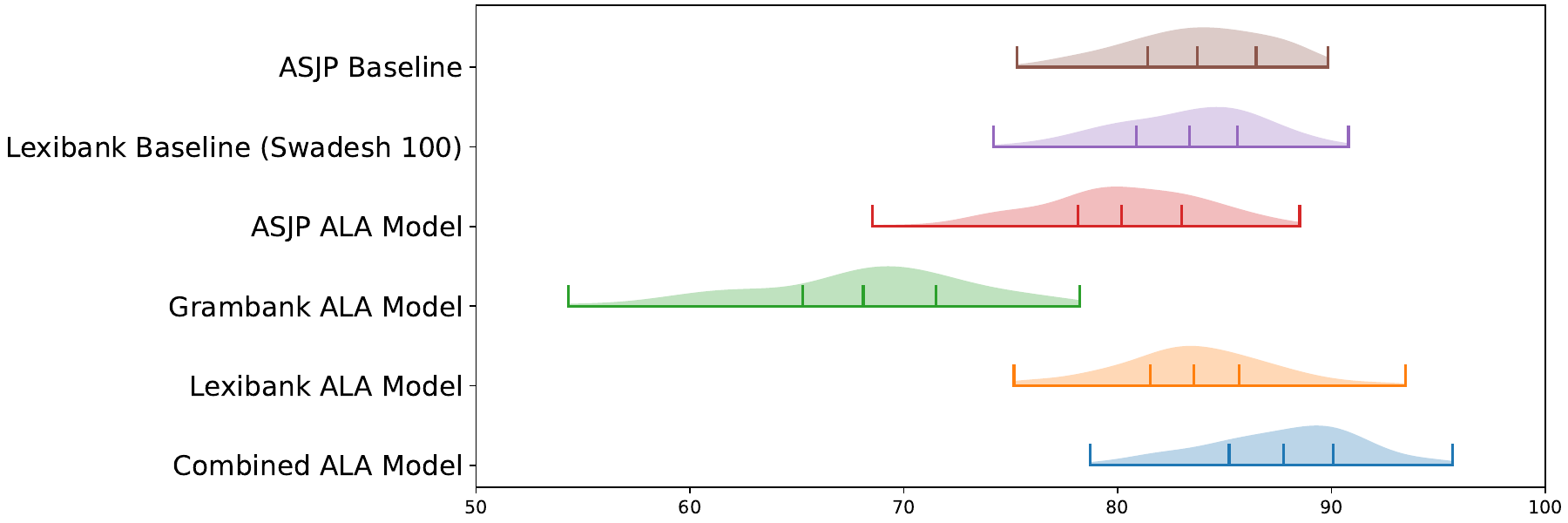}
        \caption{Classification results for all models, based on 100 runs with random seeds for the train/test split. Vertical lines indicate minimum, maximum, 25th, and 75th percentile, as well as the mean. The comparison is based on the balanced accuracy across language families to account for the difficulty of classifying small language families.}
        \label{fig:results}
\end{figure*}

We compared our two baseline models (ASJP and Lexibank) to four neural network models for automated language classification (ASJP, Grambank, Lexibank, and Grambank/Lexibank combined). In this test, we selected all 1057 languages common to ASJP, Lexibank, and Grambank, which belong to language families with at least five members. The classification target consisted of 29 different language families, including one for \emph{isolates} (languages not assigned to any family).

\begin{table}[t]
\centering
\tabular{|l|c|c|}
\hline
\bfseries Model & \bfseries Accuracy & \bfseries SD \\\hline\hline
ASJP Baseline & 83.74 &  3.25 \\\hline
Lexibank Baseline & 83.36 & 3.35 \\\hline
ASJP ALA &  80.13 & 3.85 \\\hline
Grambank ALA & 68.11 & 5.07 \\\hline
Lexibank ALA & 83.73 & 3.64 \\\hline
Combined ALA & 87.75 & 3.59 \\\hline
\endtabular
\caption{Results of all models in the model comparison.}
\label{tab:res}
\end{table}

The results in Figure \ref{fig:results} and Table \ref{tab:res} show the performance of all six models, based on the balanced accuracies from each of the 100 runs. The lexical models strongly outperform the model relying exclusively on grammatical data. The baseline models perform similarly to the Lexibank neural network model, while the neural network model using ASJP data falls off. The combined Lexibank/Grambank neural network model outperforms all models by about four points in accuracy.

Our results show that the simple baseline models perform on par with the more complex model structures. The combined model is the only exception, outperforming all other models in the comparison. This suggests that language affiliation benefits from a holistic approach combining lexicon and grammar data, confirming traditional assumptions from historical linguists. To further explore the potential of neural network approaches to automated language affiliation, we concentrate on the models based on Grambank, Lexibank, and their combined data in the following experiments. 
\subsection{Finding Deep Genealogical Relations}\label{test2}

\subsubsection{Indo-European}
We conducted three case studies testing the affiliation of entire subgroups. Our first test is based on the Indo-European language family, spoken mainly in Europe, Northern India, and the Iranian plateau. The time depth for the initial split of the first branches, Anatolian and Tocharian, from the rest of the language family, is contested, with individual proposals ranging from 6,000 years ago \cite{Anthony2015} up to 8,000 years before present \cite{Heggarty2023}. We separate the languages from those two branches from the training data. The Lexibank model correctly classifies the languages as Indo-European (100\%). Additional tests could not be carried out, since the languages in question are not coded for Grambank.

\subsubsection{Sino-Tibetan}
Estimates for the age of the Sino-Tibetan language family range between 5,900 years \cite{Zhang2019} and 7,200 years \cite{Sagart2019}. In our test, we separate the languages of the Sinitic branch (the Chinese languages), commonly believed to be one of the earliest branches to split off from the ancestral proto-language, from the training data, and train our models without them.
Similar to the test on Indo-European languages, the Lexibank model successfully classifies the Sinitic branch to be part of the Sino-Tibetan language family, with an overall accuracy of 87.5\%. The combined model surpasses the Lexibank model in accuracy, reaching 98\%.
The Grambank model classifies only one variety (\href{https://glottolog.org/resource/languoid/id/wutu1241}{wutu1241}) primarily as Sino-Tibetan, whereas the other varieties from Sinitic tend to be classified either as Hmong-Mien (\href{https://glottolog.org/resource/languoid/id/mand1415}{mand1415}, \href{https://glottolog.org/resource/languoid/id/wuch1236}{wuch1236}) or Austroasiatic (\href{https://glottolog.org/resource/languoid/id/hakk1236}{hakk1236}). At least in the case of Hmong-Mien, the classification of the grammatical model points to an important role of areality.

\subsubsection{Uto-Aztecan}
Uto-Aztecan is one of the largest language families spoken in North America, located primarily on the west coast of the Pacific \cite{Campbell1997}. It consists of two main branches, the northern and the southern languages. This split is estimated to have occurred between 3,258 and 5,025 years ago \cite{Greenhill2023}. The Lexibank model classifies the northern branch as Uto-Aztecan in about 40\% of cases. The grammatical model fails in this classification task and only achieves 10\% accuracy. Instead, the languages are classified as Cariban, Chibchan, or Pama-Nyungan. In this case, the grammatical data also seems to drag down the accuracy of the combined model (26\%), which performs worse than the Lexibank data alone.

\subsection{Affiliation of Isolates}\label{test3}
\begin{figure*}[t]
    \centering
    \includegraphics[width=0.9\textwidth]{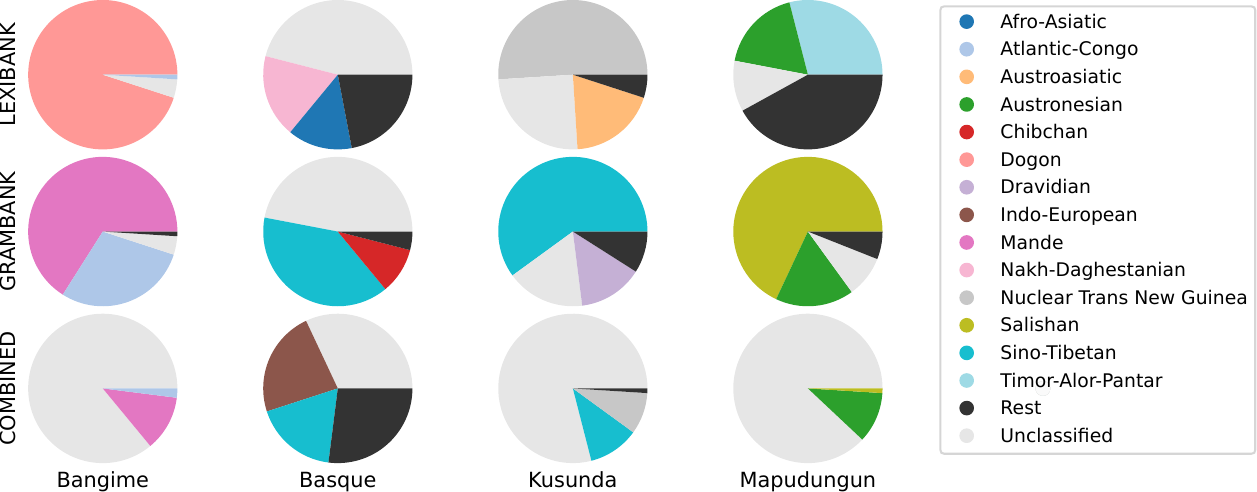}
    \caption{Results for the experiment on isolate affiliation. Results are limited to the first three families to which an isolate is affiliated, showing the proportion of the remaining families under the label \textit{Rest} in the charts.}
    \label{fig:isolates}
\end{figure*}

We test our models on four language isolates that have found recent attention in the literature. Bangime, a language spoken in central-eastern Mali, is considered an isolate that has been in contact with many surrounding languages for such a long time that the speakers consider it related to the neighboring Dogon languages \cite{Hantgan2022}. The data on Bangime in Lexibank is based on \citet{Hantgan2022}. Basque has been recently hypothesized to be part of a Proto-Euskarian-Indo-European language family using both traditional and computational methodology \citep{Blevins2018, Blevins2021}. The data on Basque in Lexibank is taken from \citep{Dellert2020}. Kusunda is spoken in Nepal, but has previously been hypothesized to be related to Papuan languages \cite{Whitehouse2004}, the Dene family, or Yenisseien \cite{Gerber2017, vanDriem2014}. However, no classification has found broad acceptance, and the language remains classified as isolate. A new wordlist recorded in 2020 by \citet{Aaley2020} has been included in Lexibank. Finally, Mapudungun is a language spoken in south-eastern South America. The available evidence suggests that Mapudungun's genealogical relations with other languages are restricted to close relatives that have since become dormant, with no indication of deeper connections. The data for Mapudungun in Lexibank is based on \citet{Tadmor2009} with phonetic mappings by \citet{Miller2020}.

The results are presented in Figure~\ref{fig:isolates}. Bangime is classified consistently as Dogon in the Lexibank model (95\%) and as Mande (66\%) or Atlantic-Congo (29\%) in the Grambank model. Consequently, the combined model primarily has Bangime unclassified (86\%). Basque on the other hand remains mostly unclassified in both the Lexibank (46\%) and the Grambank model (47\%), although the latter also tends to propose an affiliation with Sino-Tibetan (39\%). The combined model proposes an affiliation with Indo-European (23\%) or Sino-Tibetan (18\%) but also includes the unclassified affiliation (32\%). In the Lexibank model, Kusunda is affiliated either with Nuclear Trans-New Guinea (51\%), Austroasiatic (19\%), or left unclassified (17\%). The Grambank model mostly affiliates Kusunda with the Sino-Tibetan family (60\%). The combined model mostly proposes no affiliation of Kusunda with other language families (79\%). Mapudungun is split between several language families in the Lexibank model: Timor-Alor-Pantar (29\%), Austronesian (18\%), or Unclassified (11\%). The Grambank model mostly suggests a Salishan affiliation (68\%) or no affiliation (17\%). The combined model, again, finds no clear affiliation pattern (88\%).

Given that the status of all four languages concerning their affiliation with other language families has been disputed without a result for a long time, it would go too far to speculate on any particular finding presented in the charts here. What we can see, however, is a tendency for the combined model to affiliate the four isolates with the large group of unclassified (i.e., isolate) languages in our sample. The Lexibank and Grambank models differ quite remarkably in this regard, often giving preference to particular language families. 

The Lexibank model classifies Bangime as a Dogon language, reflecting the well-known fact that the lexicon shares many words with this family \citep{Hantgan2022b}, whereas the grammatical model suggests an affiliation with Mande languages. If we consider the results from the previous tests indicative and the affiliation of the grammatical model as being prone to contact phenomena, heavy grammatical restructuring for Bangime based on Mande languages seems likely. This mirrors previous arguments that typological features reflect geographical distributions, rather than genealogical relations \cite{Greenhill2010, Gray2010, Donohue2011}, and contributes directly to our understanding of the linguistic layers of Bangime, which so far focused on the lexicon \cite{Hantgan2022}. Given that genetic studies show that the speakers are unique in the region concerning their genetic history \cite{Babiker2020}, it is reasonable to classify Bangime as an isolate, as does the Combined model. 

For Kusunda, we find a lexical link to Nuclear Trans-New Guinea languages, while the grammar fits well into the Sino-Tibetan neighborhood where the language is spoken. This finding aligns with previous research that suggests that Kusunda is genealogically a Trans-New Guinea language that has recently migrated to its current location \cite{Whitehouse2004, vanDriem2014}. This is where Kusunda would have come into contact with Sino-Tibetan languages, adopting several grammatical features from this language family.

The closeness between Basque and Indo-European suggested by the combined model has been recently suggested using both traditional and computational methods \cite{Blevins2018, Blevins2021}. The connection of Basque with Sino-Tibetan suggested by the Grambank model and the connection with Nakh-Daghestanian suggested by the Lexibank model would fit with the far-ranging proposal of a Sino-Caucasian macro-family, in which scholars at times include Basque \citep{Starostin2017}.

For Mapudungun, few prominent classification hypotheses exist in the literature \cite{Campbell2012}. All our models tend to affiliate the language to some degree with Austronesian, with some suggesting a Timor-Alor-Pantar (lexical) or Salishan (grammatical) affiliation. We are not aware of previous mentions of those affiliations. One way to explore them would be to analyze the individual shared data points to evaluate the possibility of chance similarities. This points also to the major drawbacks of the neural network approach, as it does not allow us to directly determine the concrete words or grammatical features that contribute to a particular decision.


\subsection{Classification of Unaffiliated Languages}\label{test4}
\begin{figure*}[t!]
    \centering
    \resizebox{\textwidth}{!}{%
    \tabular{cc}
        \includegraphics[height=1.25cm]{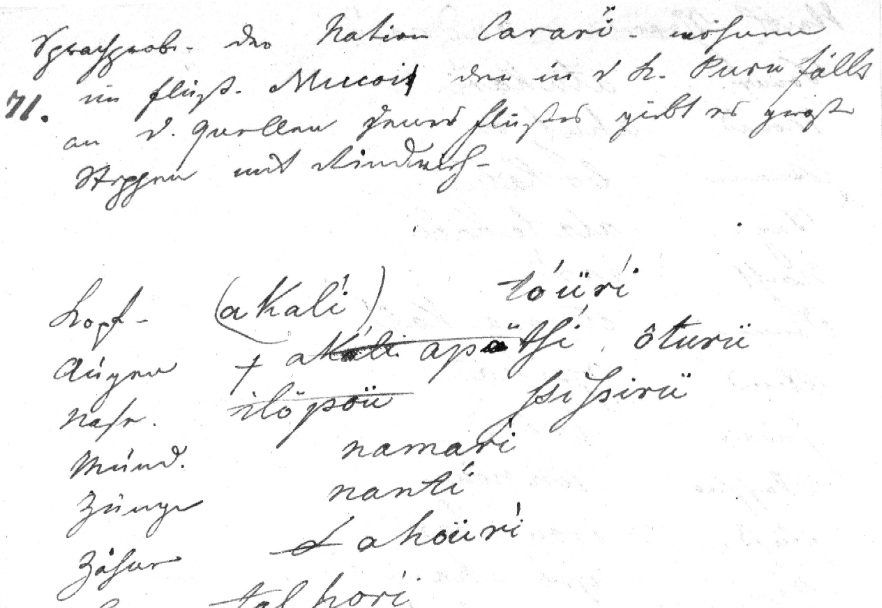}
        &
        \includegraphics[height=1.25cm]{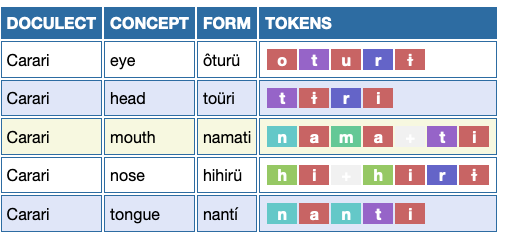}
    \endtabular}
    \caption{The left shows some of the original Cararí data published by \citet{Natterer}. The right shows our standardization of the same entries using the EDICTOR tool \cite{EDICTOR-3.1}, with the original transcriptions given in the column `Form'.}
    \label{fig:carari}
\end{figure*}

The lexical model can also affiliate newly identified or historical languages documented in ancient sources to language families. To illustrate this, we affiliated data from Cararí, a language documented during the early 19th century by Johann Natterer at the confluence of the Mucuim River and the Purús in the Brazilian Amazon \cite{Natterer}. According to \citet{Adelaar2014}, some of the languages documented by \citeauthor{Natterer} could not yet be classified -- Cararí being one of them. However, they suggest that an Arawak affiliation might be identifiable with more comparative work done in the future.

In the first step, we digitized the data and converted the wordlist into the Lexibank format. Figure~\ref{fig:carari} presents the original and the standardized versions of the documented word forms. Through this formatting, we can easily input this data into the lexical model by adding the database with the same workflow as the rest of the data. The results are clear: the Lexibank model strongly suggests an Arawak affiliation of the language (80\%).

\section{Discussion and Conclusion}
We have presented in this study a new approach to \textit{automated language affiliation}. The lexical and combined models show good classification results despite their simple architecture. Given the strong performance, we can use the models for downstream tasks such as testing long-distance relationships between subgroups of language families and the affiliation of unclassified language varieties. The first experiment shows that our models correctly identify such long-distance relationships at a time depth of 5,000 years and more. When testing the affiliation of linguistic isolates, our models reflect the actual discussions in the linguistic literature. This shows that our method can extract information from sparse data in a way that can be compared with language-specific studies.

From our findings, we can make three conclusions, (a) language affiliation achieves promising results even for language relations way back in time, (b) grammar alone is not sufficient for a successful affiliation, and (c) combined models seem to work very well, reflecting that languages are best affiliated by using lexicon plus a bit of grammar \cite{Campbell2008}. However, the combined data is only available for a small subset of languages. In cases of data scarcity, the lexical models are almost on par with the combined model and strongly outperform comparable models based on grammatical data \cite{Holman2008}. In individual cases (e.g. testing Uto-Aztecan), the lexical model even outperforms the combined model. 


A particular use case of our method is the affiliation of historical data with contemporary language families. In many cases, the material is so scarce that cannot be affiliated with any language family based on a traditional analysis alone. Our models provide a quantitative perspective, and the case of Cararí shows that it might even be possible to provide strong arguments for a specific affiliation.

We do not see automated language affiliation replacing the traditional comparative method. While our model of language affiliation can be used to evaluate hypotheses about long-range genealogical relationships between languages, it cannot provide conclusive proof in favor or against such relationships. The strength of our approach is a principled comparison of the data across a large range of languages that can find hints at a shared descent between languages. This can be a starting point for a linguistic evaluation of such hypotheses, which would be verified, for example, through the means of cognate reflex prediction \cite{Blum2024d} or other traditional workflows \cite{Durie1996}. The task of evaluating those relations will have to remain with the comparative method, which could now target specific proposals to shed further light on the history of human languages.

\section*{Data and Code Availability}
All data and code needed to replicate this study along with detailed instructions can be accessed from the Open Science Framework via the following link: \url{https://osf.io/wqt2j/?view_only=016a4e833dec445ebc4341fdf0e23f37}.

\section*{Competing Interests}
The authors declare no competing interests.

\section*{Limitations}
Data for many small language families is scarce. Even though we use datasets with data from more than 2,000 languages, they only represent about a third of the world's language families. All models showed that classification is much more difficult for small language families. Considering the availability of training data in those settings, this is expected. While projects like ASJP \cite{ASJPv20} opt for smaller concept lists from more languages due to this reason, we are convinced that there are several advantages of using the explicit orthography conversion from Lexibank, even though this means a trade-off in terms of languages available.

This also influences the comparability of our models. Only a subset of around 1,000 languages is available in all three major datasets. We only use this set of common languages to compare all models using the same data. Ideally, we would have a larger subset with a better representation of the worldwide linguistic diversity.

\section*{Ethical Considerations}
We see no potential risks involved, except for the potential of creating unwarranted hypotheses about long-distance genealogical relationships between languages. We call for all users of our models to sensibly analyze the linguistic data behind the proposed classifications.

\bibliography{acl_latex}

\end{document}